\documentclass[11pt]{article}

\usepackage[final]{acl}

\usepackage{times}
\usepackage{latexsym}

\usepackage[T1]{fontenc}

\usepackage[utf8]{inputenc}

\usepackage{microtype}

\usepackage{inconsolata}

\usepackage{graphicx}
\usepackage{booktabs}
\usepackage{amsmath}
\usepackage{multirow}
\usepackage{hyperref}
\usepackage{url}
\usepackage{graphicx}
\usepackage{booktabs}
\usepackage{subcaption}
\usepackage{tcolorbox}
\tcbuselibrary{breakable}
\usepackage{amsfonts}

\title{MetaEvo: A Meta-Optimization Framework for Experience-Driven Agent Evolution}

\author{
 \textbf{Bowen Ren\textsuperscript{1}},
 \textbf{Heyan Huang\textsuperscript{1,2}},
 \textbf{Yinghao Li\textsuperscript{1}},
 \textbf{Yang Gao\textsuperscript{1,2}},
\\
\\
 \textsuperscript{1}School of Computer Science and Technology, Beijing Institute of Technology, Beijing, China \\
 \textsuperscript{2}Beijing Institute of Technology Southeast Academy of Information Technology, Putian, China
\\
 \texttt{\{bwren-bit, hhy63, yhli, gyang\}@bit.edu.cn} \\\
}

\begin{document}
\maketitle
\begin{abstract}
Large language models (LLMs) exhibit strong reasoning capabilities, yet most LLM-based agents are statically deployed and unable to improve through task interactions. Existing experience-driven methods often rely on memory or heuristics without enhancing the model’s ability to learn, treating it as a passive executor and leading to early performance plateaus and limited long-term improvement.
To address this issue, we propose MetaEvo, a two-stage framework for continual agent evolution that focuses on improving how the model learns from tasks experience, rather than solely on what it stores. MetaEvo first applies preference-based optimization to enhance the model’s ability of principle abstraction, then enables the accumulation and reuse of these principles within a modular agent architecture.
Experimental results on diverse reasoning benchmarks demonstrate that MetaEvo consistently outperforms strong baselines, maintains reliable improvement across iterations. These findings validate the effectiveness of meta-optimization in enabling agents to learn from experience and continually enhance their reasoning capabilities.
\end{abstract}

\section{Introduction}
\begin{figure}
    \centering
    \includegraphics[width=1\linewidth]{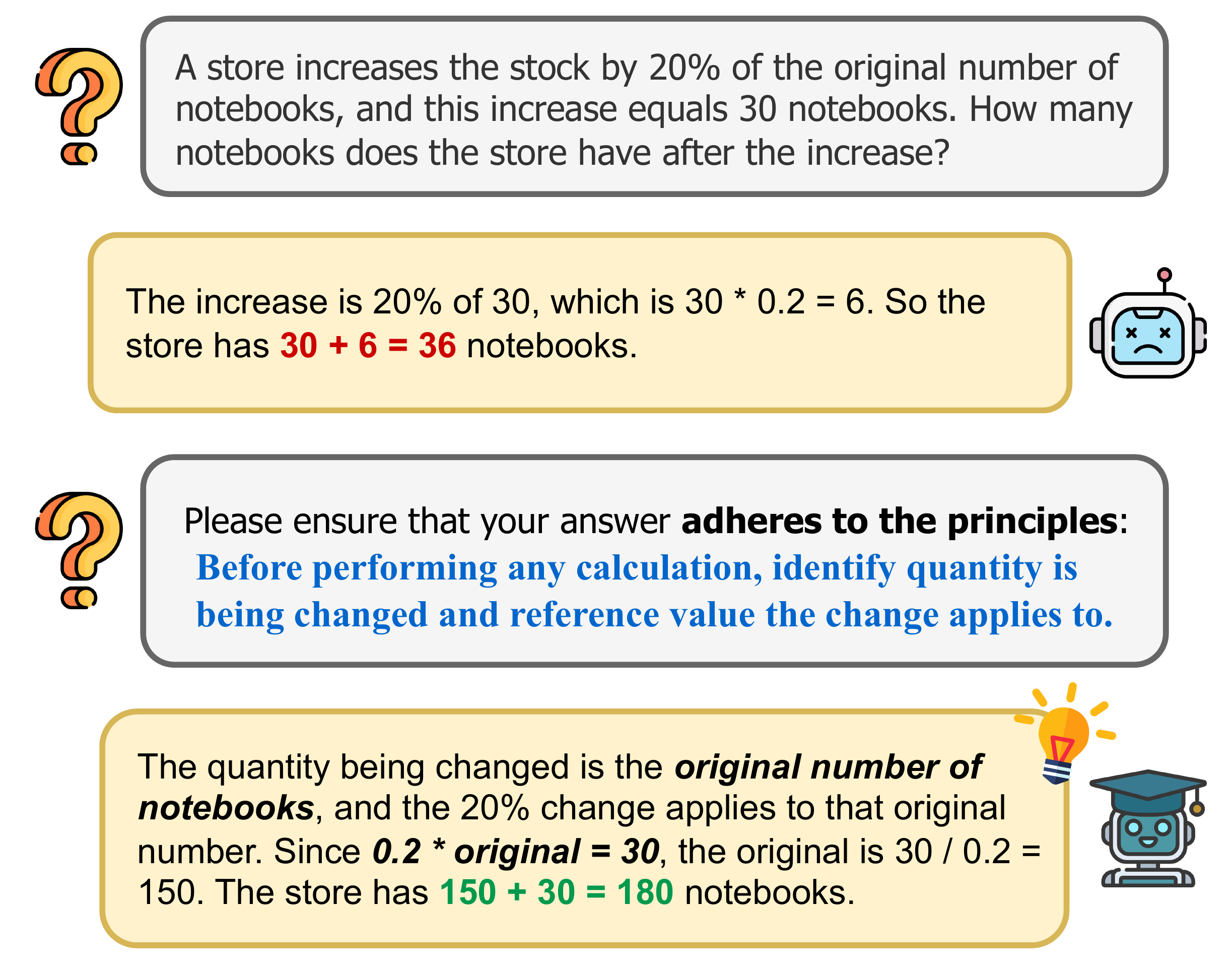}
    \caption{An instance of principle guided generation.}
    \label{fig:principle_example}
\end{figure}

Large Language Models (LLMs) have demonstrated strong performance across a wide range of natural language processing tasks~\citep{DBLP:journals/corr/abs-2005-14165, DBLP:journals/corr/abs-2302-13971, c:22}.
However, most LLM-based agents are statically deployed and cannot accumulate or reuse knowledge learned from past successes and failures, collectively known as \textbf{experience},  leading to repeated reasoning or planning errors~\citep{DBLP:conf/nips/MadaanTGHGW0DPY23, DBLP:conf/nips/ShinnCGNY23, li2023largelanguagemodelsunderstand, DBLP:conf/iclr/GouSGSYDC24, DBLP:conf/emnlp/YangLL23, DBLP:journals/corr/abs-2401-03385}. 
Recent research has begun to explore experience-driven agent evolution by distilling accumulated task interaction data into structured and memory-based knowledge, enabling agents to progressively refine their behavior and support continual self-evolution. As illustrated in Figure~\ref{fig:principle_example}, many existing methods represent accumulated task experience as high-level textual \textbf{principles} that provide guidelines for correcting model behavior and are explicitly injected into the context during inference to guide reasoning and decision-making~\citep{DBLP:conf/acl/GaoDCZW0024, DBLP:conf/aaai/Zhao0XLLH24, cai2025buildingselfevolvingagentsexperiencedriven}.

While such principles provide an effective way to summarize past mistakes and guide future behavior correction, they are not continually optimized and exhibit unstable quality, positioning the LLM primarily as a passive executor rather than an active adapter during agent evolution. The fundamental limitation is that existing approaches treat learning from experience as a static procedure, lacking a mechanism to optimize the capacity for learning itself, which often leads to limited improvement during evolution and prevents sustained performance gains.

To address this issue, we treat the process of principle extraction as a learnable and optimizable capability, framing it as an evolving meta-ability. To achieve this, we propose MetaEvo, a framework that shifts the focus from optimizing performance outcomes to dynamically enhancing the learning process. The framework consists of two stages: meta-optimization for improving principle abstraction, and principle accumulation, an iterative evolutionary cycle.

Specifically, in the first stage, we enhance the model’s meta-ability to learn how to abstract high-quality principles by distilling insights from stronger and more abstract alternatives, referred to as meta-optimization. To achieve this, we leverage a more capable external model to construct a preference dataset and then apply Direct Preference Optimization (DPO)~\cite{rafailov2023direct} to align the model’s outputs with the preferred principles.
In the second stage, we build a principle library using the meta-optimized model and apply it to guide generation during the inference phase. Through multiple iterations of this procedure, the principle library is progressively refined, enabling it to provide increasingly effective guidance for the model.

The framework is implemented through a modular agent system consisting of \textit{plan}, \textit{memory}, and \textit{execution} modules. Within the \textit{plan} module, to address the issue that principles derived by current methods are often overly generic and misaligned with concrete reasoning errors, we design a Contrast-Driven Abstraction (CDA) method that enables the model to generate targeted and actionable principles by contrasting fine-grained differences in answers. This module is applied in both stage of MetaEvo. Besides, the \textit{memory} module maintains a structured repository of principles, while the \textit{execution} module is responsible for retrieving and applying these principles; both modules are applied in the second stage.

We evaluate MetaEvo across diverse reasoning benchmarks and observe consistent improvements over strong baselines, validating the effectiveness of meta-ability optimization. Moreover, meta-ability enables sustained performance improvement through iterative self-evolution, rather than short-lived gains. Our contributions can be summarized as follows: 
\begin{itemize}
    \item  We propose MetaEvo, a  experience-driven framework featuring meta-optimization and principle accumulation, and evaluate its effectiveness across diverse reasoning benchmarks, where it consistently outperforms competitive baselines.
    \item We introduce meta-optimization as a learning paradigm for agent systems, showing that optimizing intermediate capabilities leads to more effective self-improvement than directly optimizing final outputs.
    \item We propose a contrast-driven principle extraction method CDA, which ensures that the derived principles directly address underlying strategic errors and provide actionable guidance.
\end{itemize}

\section{Related Work}

\subsection{Experience-Driven Evolution.}
Recent work has explored self-evolving agents that improve through continual learning from experience~\citep{gao2025surveyselfevolvingagentspath}. A major line of research focuses on identifying model failures and distilling corrective principles or reusable knowledge, which are incorporated via auxiliary supervision or external memory to guide future behavior~\citep{DBLP:conf/nips/SunSZZCCYG23, DBLP:conf/emnlp/YangLL23, DBLP:conf/nips/MadaanTGHGW0DPY23}. Related methods enhance adaptability through retrieval-augmented inference and memory-based prompting, enabling models to reference prior corrections during generation. Additional efforts address scalability via structured experience replay and memory management~\citep{DBLP:conf/acl/GaoDCZW0024, DBLP:conf/aaai/Zhao0XLLH24, DBLP:conf/naacl/GongHMNDZTFGV24, liu2025contextualexperiencereplayselfimprovement, ouyang2025reasoningbankscalingagentselfevolving, xu2025sedmscalableselfevolvingdistributed}. Experience-driven evolution has also been extended to embodied agents that refine actions through real-world feedback~\citep{li2024robocoderroboticlearningbasic}.

Separately, meta-level optimization has gained increasing attention, where the learning process itself becomes the object of improvement. Some approaches introduce secondary evaluators to shape learning rewards~\citep{xiong2025mpoboostingllmagents}, while others design higher-order planning mechanisms that reason over planning strategies~\citep{wu2025meta}. These methods share MetaEvo’s high-level goal of enhancing model performance by optimizing beyond direct task execution, MetaEvo enables the model to learn how to derive better principles for future self-improvement.

\subsection{Memory-based Agent System}
Equipping LLM agents with external memory has emerged as a core approach for enabling continual learning, behavioral adaptation, and long-horizon reasoning.
Early studies primarily rely on experience replay, storing intermediate reasoning steps or error-feedback pairs to guide future decisions~\citep{DBLP:conf/emnlp/YangLL23, DBLP:conf/emnlp/LiQ23, DBLP:conf/acl/GaoDCZW0024}. 
Complementary to this line, structural memory methods organize stored information into semantically or functionally structured segments, improving interpretability and retrieval efficiency~\citep{zeng2024structuralmemoryllmagents, DBLP:conf/aaai/Zhao0XLLH24}.
More recent work shifts attention to memory management and evolution for long-term agent behavior. 
MemoryBank~\citep{zhong2023memorybankenhancinglargelanguage} and Task-Core Memory~\citep{huai2025taskcorememorymanagementconsolidation} study retention and consolidation mechanisms to mitigate forgetting in continual settings. 
Beyond static storage, A-MEM~\citep{xu2025amemagenticmemoryllm} and Mem0~\citep{chhikara2025mem0buildingproductionreadyai} explicitly model memory evolution, enabling agents to refine and reorganize stored knowledge over time. 

Collectively, these mechanisms represent a paradigm shift from static retrieval-based memory to dynamically evolving knowledge architectures.

\begin{figure*}[ht]
\centering
\includegraphics[width=1\linewidth]{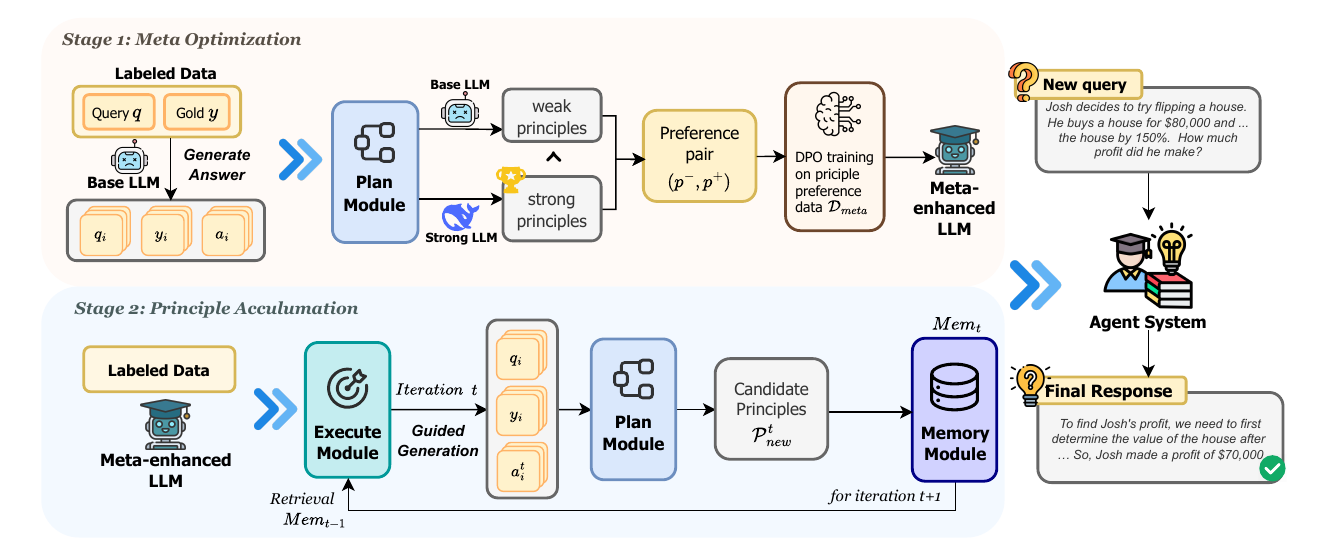}
\caption{
Illustration of the pipeline of the \textsc{MetaEvo} framework. (1) \textbf{Meta Optimization:} We first train a model to enhance its core meta-ability through preference-based learning on principles. (2) \textbf{Principle Accumulation:} The enhanced model then abstracts and accumulates a refined set of principles into a structured \textit{memory} module, which can be iteratively expanded. At inference time, the agent retrieves the most relevant principles from memory to steer its final response.
}
\label{fig:framework}
\end{figure*}

\section{Methodology}

This section presents MetaEvo, a meta-optimization framework implemented as a modular agent system, that enables experience-driven and principle-guided evolution through its three core modules: \textit{plan}, \textit{memory}, and \textit{execution}.
Accordingly, we first present the overall workflow of the framework, and then delve into the specifics of each constituent module.

\subsection{Framework Pipeline}

\subsubsection{Meta-Optimization}

In this stage, we fine-tune the base LLM to enhance its \textbf{meta-ability}, the capacity to learn how to abstract and internalize actionable and instructive principles from experience. Rather than learning to solve tasks directly, the model is optimized to develop a preference for principles that provide clearer and more operational corrective guidance, formulated as a meta-optimization problem over revision knowledge.

For each query $q_i$, the base model first produces an initial answer $a_i$. A principle abstraction process then takes $(q_i, a_i, y_i)$ to derive a revision principle. We perform this abstraction twice: once using the base model to obtain a less instructive principle $p_i^-$, and once using a more capable external LLM to obtain a preferred principle $p_i^+$. This yields a preference pair $(p_i^+, p_i^-)$ for each query, forming the meta-optimization dataset:
\begin{equation}
\mathcal{D}_{\text{meta}} = \left\{ \left( q_i, p_i^+, p_i^- \right) \right\}_{i=1}^N .
\end{equation}

We optimize the base model on $\mathcal{D}_{\text{meta}}$ using Direct Preference Optimization (DPO), which incorporates pairwise preference supervision by minimizing the expected loss:
\begin{equation}
\min_{\theta} \, \mathbb{E}_{(q, p^+, p^-) \sim \mathcal{D}_{\text{meta}}}
\left[ \mathcal{L}_{\text{meta}}(\pi_\theta; q, p^+, p^-) \right],
\end{equation}
where $\pi_\theta$ denotes the model parameterized by $\theta$. The loss encourages higher likelihood for the preferred principle $p^+$ over the dispreferred $p^-$:
\begin{equation} \begin{split} \mathcal{L}_{\text{meta}} &= -\mathbb{E}_{(q, p^+, p^-)} \bigg[ \log \sigma \Big( \beta \big( \\ & \quad \log \pi_\theta(p^+ \mid q) - \log \pi_\theta(p^- \mid q) \big) \Big) \bigg] \end{split} \end{equation}
where $\sigma(\cdot)$ is the sigmoid function and $\beta$ is a temperature parameter.

This process yields a \textbf{meta-enhanced} LLM with an improved capacity for abstracting and internalizing corrective principles, serving as the foundation for the subsequent principle accumulation stage.

\subsubsection{Principle Accumulation}

The objective of this stage is to leverage the meta-optimized model to construct a rich and structured repository of high-quality principles. Using the same labeled dataset, the \textit{execution} module retrieves task-relevant principles from memory to guide answer generation, producing a collection of queries, gold answers, and model-generated responses. The \textit{plan} module then extracts candidate principles from these outputs.

The extracted principles are systematically organized and stored in the \textit{memory} module, typically indexed by semantic representations of their corresponding tasks, thereby forming a task-oriented knowledge base.

This process is inherently iterative. In the first iteration ($t=1$), the memory is empty, and generation proceeds without guidance. At iteration $t$, the principle library $\text{Mem}{t-1}$ from the previous iteration is used for retrieval. Newly generated principles $\mathcal{P}^t{\text{new}}$ are then consolidated with $\text{Mem}_{t-1}$ to produce the updated memory $\text{Mem}_t$, which serves as the retrieval source for iteration $t+1$.

Benefiting from meta-optimization, the model progressively identifies previously overlooked issues and generates corrective principles at each iteration. As this process repeats, the memory expands and the model’s capabilities steadily improve, enabling continual evolution.

Upon completion of the Principle Accumulation phase, the agent is equipped with a structured memory of reasoning principles. For a new input, the agent retrieves the most relevant principles and integrates them into the model context as actionable guidance to steer the final generation.

\subsection{Agent System Components}

\subsubsection{Plan Module}

The \textit{plan} module serves as the central reasoning component of the agent system, responsible for identifying deficiencies in a given response and abstracting generalizable principles for improvement. We achieve this through a systematic method termed \textit{Contrast-Driven Abstraction} (CDA).

\begin{figure*}[ht]
    \centering
    \includegraphics[width=0.9\linewidth]{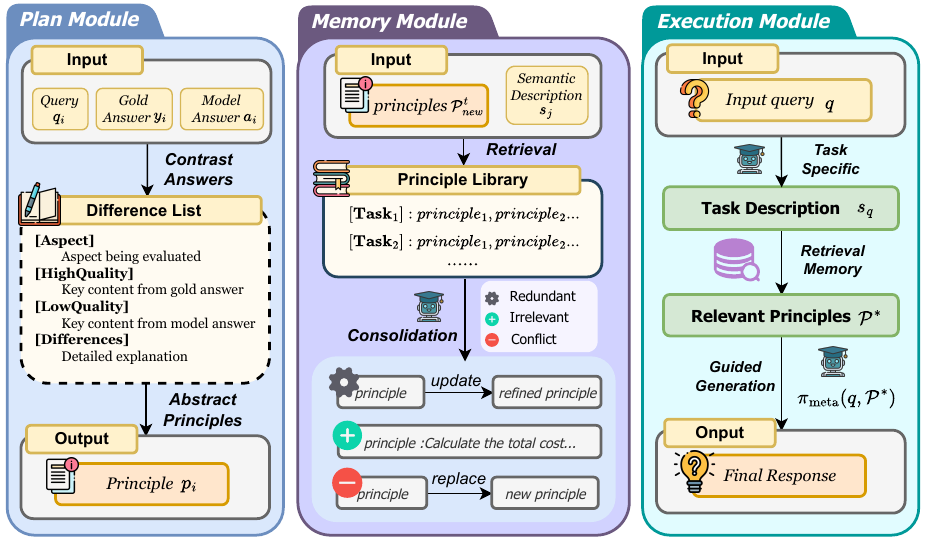}
    \caption{Overview of the agent architecture in MetaEvo.  The Plan Module derives revision principles via contrastive difference analysis. The Memory Module stores principles in a task-oriented structure and consolidates newly generated principles using three integration strategies. The Execution Module retrieves task-relevant principles to guide generation.}
    \label{fig:plan_module}
\end{figure*}

CDA operates through a two-step pipeline: (1) Discrepancy Analysis and (2) Principle Abstraction, as illustrated in Figure~\ref{fig:plan_module}.

In the first step, \textit{Discrepancy Analysis}, the LLM is prompted with the user query $q$, the base model’s response $x$, and an expert reference answer $y$. The goal of this stage is to conduct a fine-grained comparative analysis and produce a structured discrepancy representation, denoted as $\Delta$.

The resulting structure $\Delta$ enumerates the identified discrepancies as a list of entries, each comprising four elements: the aspect under comparison, a high-quality excerpt from the reference answer, the corresponding deficiency in the model-generated response, and an explanation characterizing the nature of the difference.

In the second step, \textit{Principle Abstraction}, the structured discrepancy representation $\Delta$ is provided as input to the LLM for abstraction. The model synthesizes the detailed comparisons and distills them into a single, high-quality revision principle $p$. This principle is expressed as a concise and actionable natural-language directive that captures the core lesson revealed by the contrastive analysis.

\subsubsection{Memory Module}

The \textit{memory} module serves as the agent’s long-term knowledge repository, responsible for systematically storing and managing the principles generated by the \textit{plan} module. We formalize the memory $Mem$, as a key–value structure in which each key corresponds to a task descriptor $t$, and each value is a set of task-relevant principles $\mathcal{P}t$:
\begin{equation} Mem = \{ t \mapsto \mathcal{P}_{t} \mid t \in \mathcal{T} \} \end{equation}
Here, $\mathcal{T}$ denotes the set of all task descriptors. Each descriptor $t$ is a concise natural-language summary of an input query $q$, generated by an LLM to capture the query’s core intent.

To ensure both the quality and efficiency of the repository, the memory is actively curated rather than passively appended. When a new principle $p_{\text{new}}$ is generated for a task $t$, it is validated against the existing principle set $\mathcal{P}t$. This validation process, conducted by an LLM-based evaluator, examines semantic redundancy and logical consistency. For each existing principle $p \in \mathcal{P}t$, the evaluator compares the pair $(p_{\text{new}}, p)$ and applies the following update rules:
(1) \textbf{Redundant:} If $p_{\text{new}}$ is deemed a paraphrase or minor variant of an existing principle, it replaces the older version, ensuring that the memory reflects the most current reasoning.
(2) \textbf{Conflict:} If $p_{\text{new}}$ contradicts an existing principle, both $p_{\text{old}}$ and $p_{\text{new}}$ are used to guide generation, and their correctness is evaluated. The superior principle is retained, while the other is discarded.
(3) \textbf{Irrelevant:} If $p_{\text{new}}$ is determined to be non-redundant and non-conflicting, it is directly incorporated into the memory for task $t$.

This curation mechanism prevents the accumulation of redundant or erroneous information, enabling the memory to evolve into a compact, coherent, and high-quality knowledge base that supports effective principle-guided generation.

\subsubsection{Execution Module}

The \textit{execution} module is the agent's action-taking component, responsible for generating the final, principle-guided response during inference. It leverages the curated knowledge within the \textit{memory} module to inform its generation process, which unfolds in two phases: Principle Retrieval and Guided Generation.

The retrieval phase begins with a new query $q$. The module first generates its
task semantic description $s_q$, which serves as the semantic key for searching the memory. 
The key $s_q$ is then compared against all stored task descriptions
$\mathcal{S} \subset \mathcal{M}$ to identify the most semantically similar entry, denoted as $s^{*}$.
If the similarity score of the best match exceeds a predefined threshold $\tau_r$, the entire set of principles $\mathcal{P}^{\text{*}}$ associated with $s^*$ is retrieved. 

In the Guided Generation phase, the retrieved principles $\mathcal{P}^{\text{*}}$ are incorporated into the context for a generation LLM, $\pi_{\text{meta}}$. These principles act as explicit, context-aware instructions or constraints, steering the model's output. The final response is thus generated with the benefit of proven, task-relevant strategies $\pi_{\text{meta}}(q, \mathcal{P}^{*})$

By dynamically retrieving and applying relevant knowledge, the \textit{execution} module allows the agent to generalize from past experiences to new, unseen problems, ensuring its responses are not only accurate but also strategically sound.

\section{Experiments}
\subsection{Experimental Setup}

\paragraph{Datasets}

We conduct our experiments and evaluate our model on three categories of datasets:
(1) \textbf{Arithmetic Reasoning}: This category includes GSM8K~\citep{DBLP:journals/corr/abs-2110-14168}, SVAMP~\citep{DBLP:conf/naacl/PatelBG21}, and MATH~\citep{DBLP:journals/corr/abs-2103-03874}.
(2) \textbf{Knowledge Reasoning}: We select the MMLU \cite{DBLP:journals/corr/abs-2009-03300} benchmark, focusing on a subset of high-relevance tasks that the reasoning process is grounded in recognizing and applying known concepts, theories, or definitions. Specifically, we include: College Biology,  College Chemistry,  College computer Science,  College Mathematics,  College Medicine,  College Physics,  Computer Security.
(3) \textbf{Complex Reasoning}: This involves the Big-Bench Hard (BBH) subset \cite{DBLP:conf/acl/SuzgunSSGTCCLCZ23}. We focus on four reasoning-intensive tasks, namely Logical Deduction, Tracking Shuffled Objects, Reasoning about Colored Objects, and Causal Judgement. These tasks require consistent state tracking, rule-based inference, or causal chain reasoning, and their difficulty arises from the complexity of the reasoning process itself.

In the meta optimization stage, we use the training sets of GSM8K, SVAMP, and MATH to generate the DPO data for arithmetic reasoning tasks. For the knowledge and complex reasoning tasks, we utilize the CoT-Collection \cite{kim2023cot}, which is a consolidated dataset that unifies nine diverse tasks released in FLAN into the Chain-of-Thought format, to construct the DPO data.

\paragraph{Models}
We conduct experiments using two backbone language models: LLaMA 3.1-8B-Instruct~\citep{DBLP:journals/corr/abs-2407-21783} and Qwen 2.5-14B-Instruct~\citep{DBLP:journals/corr/abs-2412-15115}. These models are used as the initial agent foundation for generation, principle abstraction, and memory interaction.
We use DeepSeek-R1~\citep{deepseekai2025deepseekr1incentivizingreasoningcapability} as the strong language model for constructing preference pairs used in meta optimization.

\subsection{Baselines}
We employ the following baseline methods. (1) \textbf{Base Model}, the original model without additional optimization; (2) \textbf{Self-Refine}\citep{DBLP:conf/nips/MadaanTGHGW0DPY23}, an iterative framework that optimizes model outputs through a recursive loop of autonomous feedback and self-correction; (3) \textbf{Self-ICL}\citep{chen2023selficlzeroshotincontextlearning}, a zero-shot framework that prompts the model to generate its own pseudo-inputs and pseudo-labels, which are then prepended as few-shot demonstrations to guide the final inference;
(4) \textbf{SE-GPT}\citep{DBLP:conf/acl/GaoDCZW0024}, a  framework that develops task-specific expertise by retrieving past experiences, practicing on self-generated tasks, and inducing new strategies into a persistent memory library. (5)\textbf{Self-Discover}\cite{zhou2024selfdiscoverlargelanguagemodels}, a framework that enables LLMs to autonomously construct explicit, task-level reasoning paths by selecting, adapting, and composing atomic reasoning modules into an executable logical structure, which is then used to solve specific task instances;
(6) \textbf{MetaEvo w/o MO}: Our framework directly using base LLM to run principle accumulation without applying meta-optimization; (7) \textbf{MetaEvo w/o CDA}: Our framework directly generation principles via prompt without applying Contrast Driven Abstraction(CDA) method; (8) \textbf{MetaEvo}: The full framework combining meta optimization with CDA to enable experience-driven self-evolution.

\begin{table*}[ht]
\centering
\footnotesize
\setlength{\tabcolsep}{1pt}
    \begin{tabular}{l|cccccc|cccccc}
        \toprule
        \textbf{Method} 
        & \multicolumn{6}{c|}{\textbf{LLaMA3.1-8B-Instruct}} 
        & \multicolumn{6}{c}{\textbf{Qwen2.5-14B-Instruct}} \\
        \cmidrule(lr){2-7} \cmidrule(lr){8-13}
        & GSM8K & SVAMP & MATH & MMLU & BBH & Avg. 
        & GSM8K & SVAMP & MATH & MMLU & BBH & Avg. \\
        \midrule
        Base Model
        & 84.5 & 83.3 & 60.2 & 66.7 & 61.1 & 71.2
        & 92.2 & 91.5 & 71.1 & 76.5 & 76.1 & 81.5 \\
        \midrule
        Self-Refine
        & 84.0 & 83.2 & 60.4 & 67.3 & 62.7 & 71.5
        & 92.5 & 91.8 & 71.5 & 75.8 & 76.9 & 81.6 \\
        Self-ICL
        & 84.8 & 85.0 & 62.2 & 65.8 & 63.8 & 72.3
        & 91.6 & 90.3 & 69.4 & 74.6 & 77.6 & 80.7 \\
        SE-GPT
        & 87.6 & 87.1 & \underline{64.1} & \underline{68.5} & 63.9 & 74.2
        & 93.7 & 93.3 & \underline{72.8} & \textbf{80.2} & 78.3 & \underline{83.7} \\
        Self-Discover
        & 86.8 & 88.7 & 62.7 & 67.4 & \underline{64.3} & 74.0
        & \underline{95.8} & 93.4 & 71.7 & 76.7 & 76.9 & 82.9 \\
        \midrule
        MetaEvo
        & \textbf{94.1} & \textbf{93.8} & \textbf{66.3} & \textbf{69.1} & \textbf{70.4} & \textbf{78.7}
        & \textbf{97.1} & \textbf{95.2} & \textbf{73.6} & \underline{79.9} & \textbf{81.9} & \textbf{85.5} \\
        - w/o CDA
        & 87.7 & 81.3 & 59.3 & 67.9 & 61.6 & 71.6
        & 95.2 & 93.7 & 72.1 & 77.4 & \underline{77.3} & 83.0 \\
        - w/o MO
        & \underline{88.3} & \underline{90.1} & 63.4 & 67.3 & 62.4 & \underline{74.3}
        & 95.3 & \underline{93.8} & 72.2 & 76.2 & 76.4 & 83.2 \\
        \bottomrule
    \end{tabular}
\caption{Performance (\%) of reasoning and self-improvement methods across benchmarks.\textbf{Bold numbers} represent the best performance on each dataset, while \underline{underlined numbers} denote the second-best results.}
\label{tab:main-results}
\end{table*}

\section{Analysis}
\subsection{Main Result}

Table~\ref{tab:main-results} reports the evaluation results of our method across five benchmark datasets. Compared with the corresponding base models, our approach consistently achieves performance gains on all benchmarks.

On GSM8K and SVAMP, which emphasize numerical reasoning, MetaEvo yields substantial accuracy improvements, indicating enhanced error correction and generalization capabilities. On MATH, the framework demonstrates strong robustness in solving complex multi-step reasoning problems. Further gains on MMLU and BBH validate the generalizability of MetaEvo across diverse, knowledge-intensive tasks.

Overall, these results highlight the effectiveness of MetaEvo in improving both task-specific performance and cross-task generalization.

\begin{figure}
    \centering
    \includegraphics[width=0.9\linewidth]{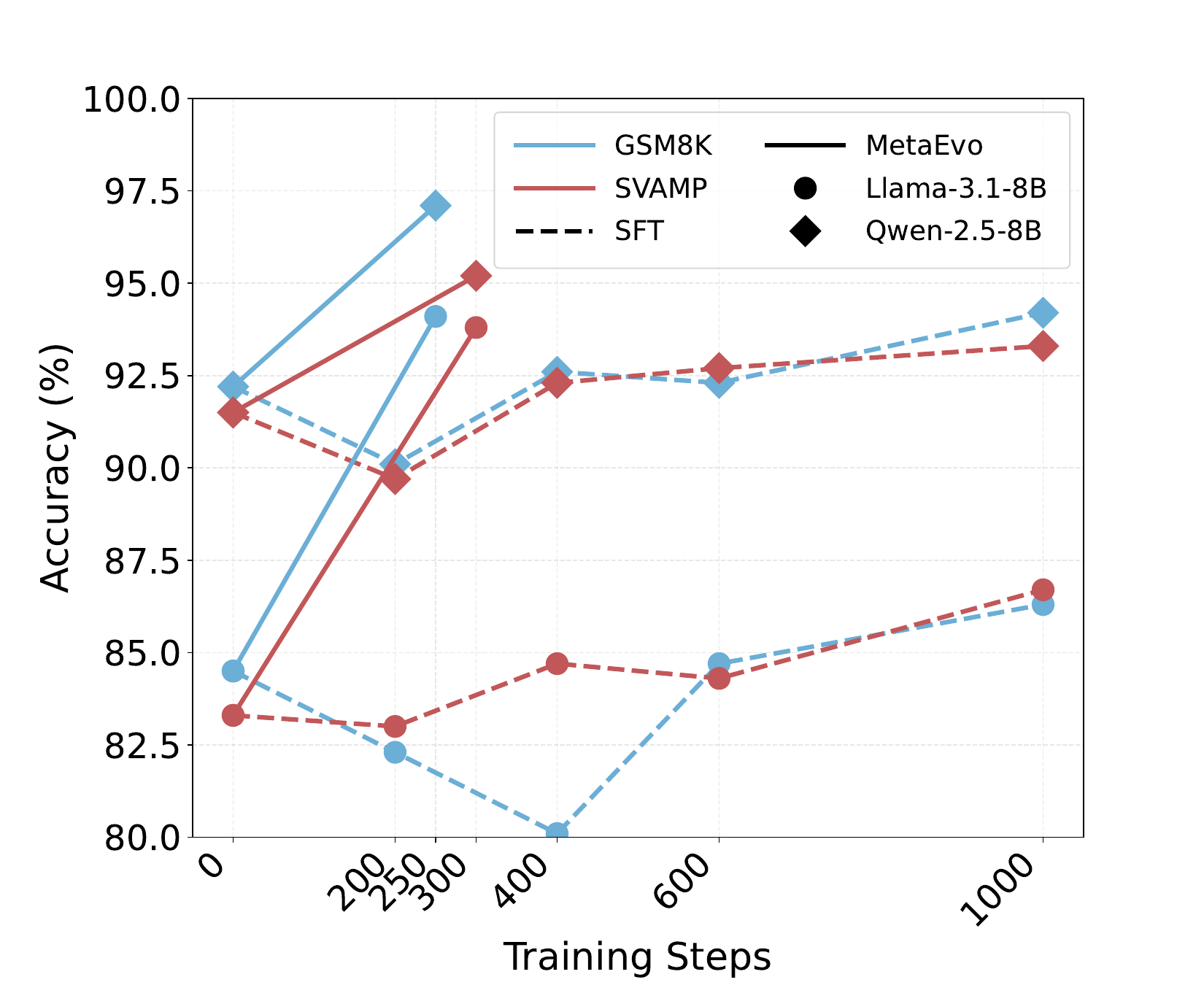}
    \caption{Compares two training methods on GSM8K. Our method achieves higher performance with fewer training samples.}
    \label{fig:methods_compare}
\end{figure}

\subsection{Meta Ability Enhances the Model’s Capacity for Self-Improvement}

\textbf{Meta optimization is critical for enabling effective principle abstraction and principle-guided generation.}
To assess the impact of meta optimization, we compare the full MetaEvo framework with a variant that excludes it (w/o MO in Table~\ref{tab:main-results}). Although w/o MO achieves competitive performance on GSM8K and SVAMP, it consistently underperforms MetaEvo across all benchmarks, highlighting the role of meta-level optimization in refining corrective principles and improving generalization.

Further analysis indicates that, without meta optimization, the extracted principles are often overly generic and lack task-specific utility, leading to redundant or inconsistent guidance. In contrast, MetaEvo produces more targeted and actionable principles aligned with observed failure modes, enabling precise revisions and effective self-correction.

Figure~\ref{fig:methods_compare} compares MetaEvo with standard supervised fine-tuning (SFT) on GSM8K and SVAMP. While SFT directly trains on labeled examples, MetaEvo learns and accumulates intermediate principles to guide generation. Despite using fewer training samples, MetaEvo consistently achieves higher performance, demonstrating the sample-efficiency gains enabled by meta optimization.

\begin{figure}
    \centering
    \includegraphics[width=0.9\linewidth]{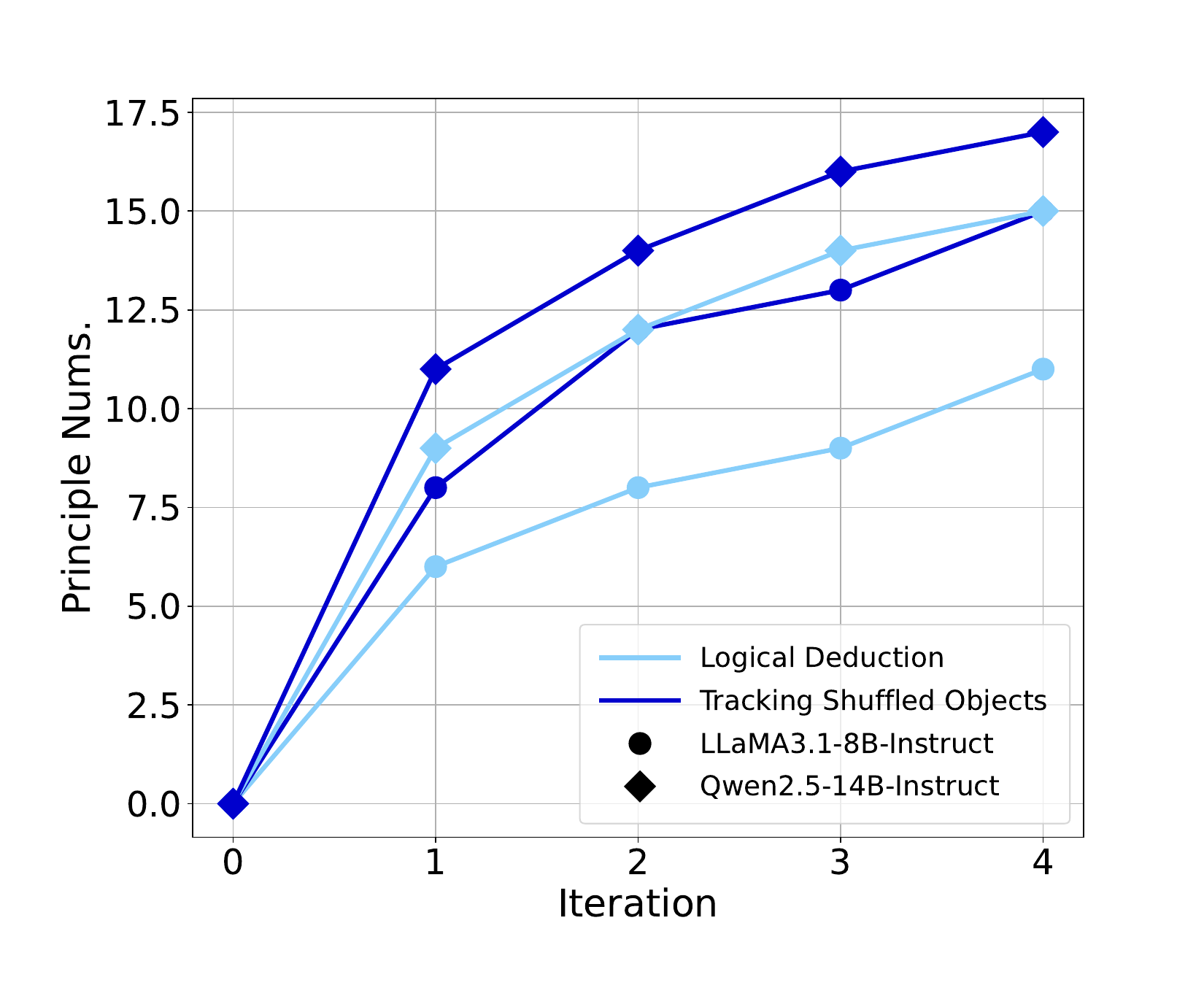}
    \caption{Principle Count Achieved Across Five Training Iterations.}
    \label{fig:principle_trend_plot}
\end{figure}

\textbf{Enhanced Meta-Ability in Principle Abstraction Raises the Performance Ceiling of Iterative Improvement.}

We further examine the effect of iterative evolution in MetaEvo by performing up to three iterations on GSM8K and SVAMP, using LLaMA3.1-8B-Instruct and Qwen2.5-14B-Instruct as base models. Each iteration consists of principle accumulation followed by principle-guided generation, where outputs from the previous iteration are reused as inputs. Across iterations, both the principle memory and response quality improve steadily.

\begin{figure}
    \centering
    \includegraphics[width=0.9\linewidth]{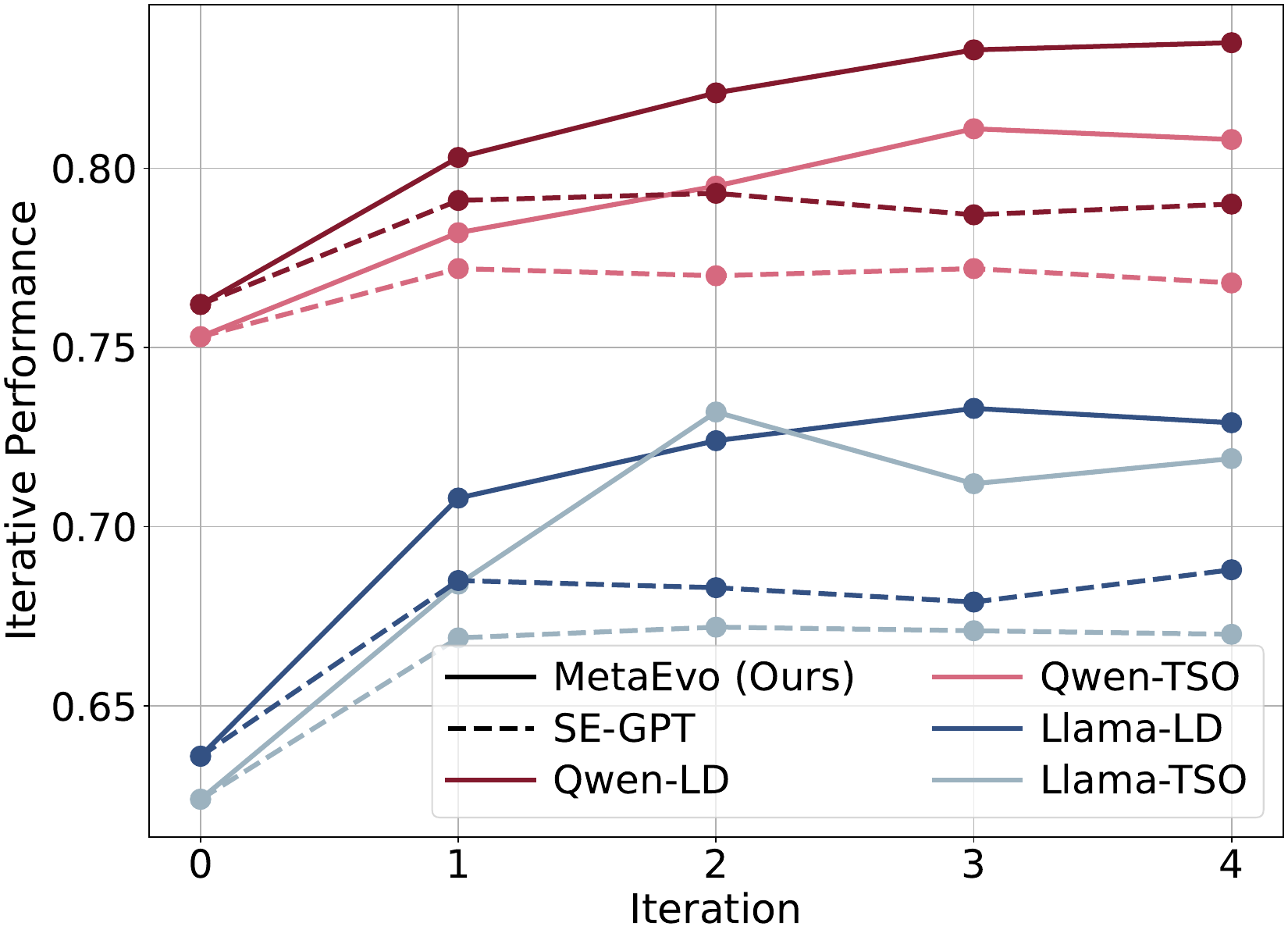}
    \caption{Accuracy(\%) comparison across iterations for MetaEvo and baseline methods. The figure reports iterative performance on two tasks of BBH: Logical Deduction (LD) and Tracking Shuffled Objects (TSO).}
    \label{fig:iterative_performance_se}
\end{figure}

As shown in Figure~\ref{fig:iterative_performance_se}, MetaEvo consistently improves across iterations and outperforms other self-evolving baselines. Performance increases monotonically over three iterations, indicating that progressive principle refinement directly enhances reasoning quality. After three iterations, MetaEvo achieves absolute gains of 9.6

These results demonstrate that meta-ability optimization enables sustained self-improvement by prioritizing the acquisition of improvement strategies over direct answer prediction.

\textbf{Meta-Optimization enhances the intrinsic reasoning ability of the model, even without explicit reuse of previous experience.} Our results demonstrate that directly strengthening the model’s capacity for principle abstraction via meta-optimization leads to consistent performance gains. As shown in Table~\ref{tab:arithmetic_results}, models trained with preference-based supervision outperform their base counterparts across multiple reasoning tasks, confirming that abstract principle alignment can improve general reasoning behavior independent of memory retrieval or prior instance reuse.

We argue that this abstraction process represents a meta-level capability that operates above specific instances, enabling the model to generalize from error patterns and revise its own behavior accordingly. Rather than memorizing task-specific corrections, the model learns to organize and apply high-level strategies that support more coherent, self-aware reasoning. This improvement feeds back into downstream performance, as the model internalizes a transferable scaffold for decision-making. In essence, principle abstraction serves not just as a mechanism for fixing past mistakes, but as a foundation for building more adaptive and generalizable reasoning behavior.

\begin{table}[ht]
    \centering
    \small
    \setlength{\tabcolsep}{5pt}
    \begin{tabular}{lcc|cc}
        \toprule
        \textbf{Method} 
        & \multicolumn{2}{c}{\textbf{LLaMA3.1-8B}} 
        & \multicolumn{2}{c}{\textbf{Qwen2.5-14B}} \\
        \cmidrule(lr){2-3} \cmidrule(lr){4-5}
        & GSM8K & SVAMP & GSM8K & SVAMP \\
        \midrule
        Base Model & 84.5 & 88.2 & 92.2 & 91.5 \\
        w/ MO      & \textbf{86.7} & \textbf{90.5} & \textbf{95.2} & \textbf{93.7} \\
        \bottomrule
    \end{tabular}
    \caption{Performance comparison with and without meta-optimization (MO) on arithmetic reasoning tasks.}
    \label{tab:arithmetic_results}
\end{table}

\begin{table}[t]
    \centering
    \small
    \setlength{\tabcolsep}{6pt}
    \begin{tabular}{lcc}
        \toprule
        \textbf{Model} & \textbf{GSM8K} & \textbf{SVAMP} \\
        \midrule
        Base Model           & 88.3 & 88.2 \\
        Random Principles    & 69.4 & 77.9 \\
        Direct Abstraction     & 88.7 & 90.5 \\
        CDA      & \textbf{92.4} & \textbf{91.9} \\
        \bottomrule
    \end{tabular}
\caption{Accuracy (\%) on GSM8K and SVAMP under different principle supervision strategies.}
\label{tab:principle-performance}
\end{table}

\subsection{Contrastive Analysis Drives Precise and Actionable Principle abstraction}
\textbf{Effective evolution relies on high-quality principles, and contrastive analysis is key to extracting effective principles.} To assess the impact of different principle generation strategies, we compare three settings: 
(1) \textit{Random Principles}, which introduce task-irrelevant noise; 
(2) \textit{Direct Abstraction}, in which principles are generated without contrastive analysis, by directly prompting the model using a predefined abstraction template; and 
(3) \textit{MetaEvo w/ CDA}, which employs contrastive-driven abstraction to derive principles.

As shown in Table~\ref{tab:principle-performance}, random principles significantly degrade performance, confirming that irrelevant guidance can mislead the model’s reasoning process. Direct abstraction produces unguided principles, which may introduce noise or even conflict with the model's original reasoning trajectory. In contrast, principles derived through contrastive analysis yield the highest and most consistent performance, achieving 92.4\% on GSM8K and 91.9\% on SVAMP, demonstrating their effectiveness in guiding generation.  These results validate the crucial role of contrastive analysis in extracting reliable, high-quality principles that serve as effective guidance for principle-guided generation.


\section{Conclusion}
In this paper, we introduce MetaEvo, a meta-optimization framework that facilitates principle-guided evolution in large language models. By enhancing the model’s meta-ability, MetaEvo shifts the objective from direct answer optimization to learning how to revise. The framework integrates meta optimization with an agent system that extracts, stores, and reuses high-quality revision principles. Experimental results across multiple reasoning benchmarks demonstrate that MetaEvo consistently improves performance, supports iterative self-improvement, and enhances generalization.

\bibliography{custom}

@misc{c:22,
      title={Attention Is All You Need}, 
      author={Ashish Vaswani and Noam Shazeer and Niki Parmar and Jakob Uszkoreit and Llion Jones and Aidan N. Gomez and Lukasz Kaiser and Illia Polosukhin},
      year={2017},
      eprint={1706.03762},
      archivePrefix={arXiv},
      primaryClass={cs.CL}
}

@article{DBLP:journals/corr/abs-2005-14165,
  author       = {Tom B. Brown and
                  Benjamin Mann and
                  Nick Ryder and
                  Melanie Subbiah and
                  Jared Kaplan and
                  Prafulla Dhariwal and
                  Arvind Neelakantan and
                  Pranav Shyam and
                  Girish Sastry and
                  Amanda Askell and
                  Sandhini Agarwal and
                  Ariel Herbert{-}Voss and
                  Gretchen Krueger and
                  Tom Henighan and
                  Rewon Child and
                  Aditya Ramesh and
                  Daniel M. Ziegler and
                  Jeffrey Wu and
                  Clemens Winter and
                  Christopher Hesse and
                  Mark Chen and
                  Eric Sigler and
                  Mateusz Litwin and
                  Scott Gray and
                  Benjamin Chess and
                  Jack Clark and
                  Christopher Berner and
                  Sam McCandlish and
                  Alec Radford and
                  Ilya Sutskever and
                  Dario Amodei},
  title        = {Language Models are Few-Shot Learners},
  journal      = {CoRR},
  volume       = {abs/2005.14165},
  year         = {2020},
  url          = {https://arxiv.org/abs/2005.14165},
  eprinttype    = {arXiv},
  eprint       = {2005.14165},
  bibsource    = {dblp computer science bibliography, https://dblp.org}
}

@article{DBLP:journals/corr/abs-2302-13971,
  author       = {Hugo Touvron and
                  Thibaut Lavril and
                  Gautier Izacard and
                  Xavier Martinet and
                  Marie{-}Anne Lachaux and
                  Timoth{\'{e}}e Lacroix and
                  Baptiste Rozi{\`{e}}re and
                  Naman Goyal and
                  Eric Hambro and
                  Faisal Azhar and
                  Aur{\'{e}}lien Rodriguez and
                  Armand Joulin and
                  Edouard Grave and
                  Guillaume Lample},
  title        = {LLaMA: Open and Efficient Foundation Language Models},
  journal      = {CoRR},
  volume       = {abs/2302.13971},
  year         = {2023},
  url          = {https://doi.org/10.48550/arXiv.2302.13971},
  doi          = {10.48550/ARXIV.2302.13971},
  eprinttype    = {arXiv},
  eprint       = {2302.13971},
  bibsource    = {dblp computer science bibliography, https://dblp.org}
}

@inproceedings{DBLP:conf/emnlp/YangLL23,
  author       = {Zeyuan Yang and
                  Peng Li and
                  Yang Liu},
  editor       = {Houda Bouamor and
                  Juan Pino and
                  Kalika Bali},
  title        = {Failures Pave the Way: Enhancing Large Language Models through Tuning-free
                  Rule Accumulation},
  booktitle    = {Proceedings of the 2023 Conference on Empirical Methods in Natural
                  Language Processing, {EMNLP} 2023, Singapore, December 6-10, 2023},
  pages        = {1751--1777},
  publisher    = {Association for Computational Linguistics},
  year         = {2023},
  url          = {https://doi.org/10.18653/v1/2023.emnlp-main.109},
  doi          = {10.18653/V1/2023.EMNLP-MAIN.109},
  bibsource    = {dblp computer science bibliography, https://dblp.org}
}

@inproceedings{DBLP:conf/aaai/Zhao0XLLH24,
  author       = {Andrew Zhao and
                  Daniel Huang and
                  Quentin Xu and
                  Matthieu Lin and
                  Yong{-}Jin Liu and
                  Gao Huang},
  editor       = {Michael J. Wooldridge and
                  Jennifer G. Dy and
                  Sriraam Natarajan},
  title        = {ExpeL: {LLM} Agents Are Experiential Learners},
  booktitle    = {Thirty-Eighth {AAAI} Conference on Artificial Intelligence, {AAAI}
                  2024, Thirty-Sixth Conference on Innovative Applications of Artificial
                  Intelligence, {IAAI} 2024, Fourteenth Symposium on Educational Advances
                  in Artificial Intelligence, {EAAI} 2014, February 20-27, 2024, Vancouver,
                  Canada},
  pages        = {19632--19642},
  publisher    = {{AAAI} Press},
  year         = {2024},
  url          = {https://doi.org/10.1609/aaai.v38i17.29936},
  doi          = {10.1609/AAAI.V38I17.29936},
  bibsource    = {dblp computer science bibliography, https://dblp.org}
}

@inproceedings{DBLP:conf/acl/GaoDCZW0024,
  author       = {Jinglong Gao and
                  Xiao Ding and
                  Yiming Cui and
                  Jianbai Zhao and
                  Hepeng Wang and
                  Ting Liu and
                  Bing Qin},
  editor       = {Lun{-}Wei Ku and
                  Andre Martins and
                  Vivek Srikumar},
  title        = {Self-Evolving {GPT:} {A} Lifelong Autonomous Experiential Learner},
  booktitle    = {Proceedings of the 62nd Annual Meeting of the Association for Computational
                  Linguistics (Volume 1: Long Papers), {ACL} 2024, Bangkok, Thailand,
                  August 11-16, 2024},
  pages        = {6385--6432},
  publisher    = {Association for Computational Linguistics},
  year         = {2024},
  url          = {https://doi.org/10.18653/v1/2024.acl-long.346},
  doi          = {10.18653/V1/2024.ACL-LONG.346},
  bibsource    = {dblp computer science bibliography, https://dblp.org}
}

@article{DBLP:journals/corr/abs-2401-03385,
  author       = {Ding Chen and
                  Shichao Song and
                  Qingchen Yu and
                  Zhiyu Li and
                  Wenjin Wang and
                  Feiyu Xiong and
                  Bo Tang},
  title        = {Grimoire is All You Need for Enhancing Large Language Models},
  journal      = {CoRR},
  volume       = {abs/2401.03385},
  year         = {2024},
  url          = {https://doi.org/10.48550/arXiv.2401.03385},
  doi          = {10.48550/ARXIV.2401.03385},
  eprinttype    = {arXiv},
  eprint       = {2401.03385},
  bibsource    = {dblp computer science bibliography, https://dblp.org}
}

@inproceedings{DBLP:conf/emnlp/LiQ23,
  author       = {Xiaonan Li and
                  Xipeng Qiu},
  editor       = {Houda Bouamor and
                  Juan Pino and
                  Kalika Bali},
  title        = {MoT: Memory-of-Thought Enables ChatGPT to Self-Improve},
  booktitle    = {Proceedings of the 2023 Conference on Empirical Methods in Natural
                  Language Processing, {EMNLP} 2023, Singapore, December 6-10, 2023},
  pages        = {6354--6374},
  publisher    = {Association for Computational Linguistics},
  year         = {2023},
  url          = {https://doi.org/10.18653/v1/2023.emnlp-main.392},
  doi          = {10.18653/V1/2023.EMNLP-MAIN.392},
  bibsource    = {dblp computer science bibliography, https://dblp.org}
}

@inproceedings{DBLP:conf/naacl/GongHMNDZTFGV24,
  author       = {Ran Gong and
                  Qiuyuan Huang and
                  Xiaojian Ma and
                  Yusuke Noda and
                  Zane Durante and
                  Zilong Zheng and
                  Demetri Terzopoulos and
                  Li Fei{-}Fei and
                  Jianfeng Gao and
                  Hoi Vo},
  editor       = {Kevin Duh and
                  Helena G{\'{o}}mez{-}Adorno and
                  Steven Bethard},
  title        = {MindAgent: Emergent Gaming Interaction},
  booktitle    = {Findings of the Association for Computational Linguistics: {NAACL}
                  2024, Mexico City, Mexico, June 16-21, 2024},
  pages        = {3154--3183},
  publisher    = {Association for Computational Linguistics},
  year         = {2024},
  url          = {https://doi.org/10.18653/v1/2024.findings-naacl.200},
  doi          = {10.18653/V1/2024.FINDINGS-NAACL.200},
  bibsource    = {dblp computer science bibliography, https://dblp.org}
}

@inproceedings{DBLP:conf/nips/ShinnCGNY23,
  author       = {Noah Shinn and
                  Federico Cassano and
                  Ashwin Gopinath and
                  Karthik Narasimhan and
                  Shunyu Yao},
  editor       = {Alice Oh and
                  Tristan Naumann and
                  Amir Globerson and
                  Kate Saenko and
                  Moritz Hardt and
                  Sergey Levine},
  title        = {Reflexion: language agents with verbal reinforcement learning},
  booktitle    = {Advances in Neural Information Processing Systems 36: Annual Conference
                  on Neural Information Processing Systems 2023, NeurIPS 2023, New Orleans,
                  LA, USA, December 10 - 16, 2023},
  year         = {2023},
  url          = {http://papers.nips.cc/paper\_files/paper/2023/hash/1b44b878bb782e6954cd888628510e90-Abstract-Conference.html},
  bibsource    = {dblp computer science bibliography, https://dblp.org}
}

@inproceedings{DBLP:conf/nips/MadaanTGHGW0DPY23,
  author       = {Aman Madaan and
                  Niket Tandon and
                  Prakhar Gupta and
                  Skyler Hallinan and
                  Luyu Gao and
                  Sarah Wiegreffe and
                  Uri Alon and
                  Nouha Dziri and
                  Shrimai Prabhumoye and
                  Yiming Yang and
                  Shashank Gupta and
                  Bodhisattwa Prasad Majumder and
                  Katherine Hermann and
                  Sean Welleck and
                  Amir Yazdanbakhsh and
                  Peter Clark},
  editor       = {Alice Oh and
                  Tristan Naumann and
                  Amir Globerson and
                  Kate Saenko and
                  Moritz Hardt and
                  Sergey Levine},
  title        = {Self-Refine: Iterative Refinement with Self-Feedback},
  booktitle    = {Advances in Neural Information Processing Systems 36: Annual Conference
                  on Neural Information Processing Systems 2023, NeurIPS 2023, New Orleans,
                  LA, USA, December 10 - 16, 2023},
  year         = {2023},
  url          = {http://papers.nips.cc/paper\_files/paper/2023/hash/91edff07232fb1b55a505a9e9f6c0ff3-Abstract-Conference.html},
  bibsource    = {dblp computer science bibliography, https://dblp.org}
}

@inproceedings{DBLP:conf/iclr/GouSGSYDC24,
  author       = {Zhibin Gou and
                  Zhihong Shao and
                  Yeyun Gong and
                  Yelong Shen and
                  Yujiu Yang and
                  Nan Duan and
                  Weizhu Chen},
  title        = {{CRITIC:} Large Language Models Can Self-Correct with Tool-Interactive
                  Critiquing},
  booktitle    = {The Twelfth International Conference on Learning Representations,
                  {ICLR} 2024, Vienna, Austria, May 7-11, 2024},
  publisher    = {OpenReview.net},
  year         = {2024},
  url          = {https://openreview.net/forum?id=Sx038qxjek},
  bibsource    = {dblp computer science bibliography, https://dblp.org}
}

@inproceedings{DBLP:conf/nips/SunSZZCCYG23,
  author       = {Zhiqing Sun and
                  Yikang Shen and
                  Qinhong Zhou and
                  Hongxin Zhang and
                  Zhenfang Chen and
                  David D. Cox and
                  Yiming Yang and
                  Chuang Gan},
  editor       = {Alice Oh and
                  Tristan Naumann and
                  Amir Globerson and
                  Kate Saenko and
                  Moritz Hardt and
                  Sergey Levine},
  title        = {Principle-Driven Self-Alignment of Language Models from Scratch with
                  Minimal Human Supervision},
  booktitle    = {Advances in Neural Information Processing Systems 36: Annual Conference
                  on Neural Information Processing Systems 2023, NeurIPS 2023, New Orleans,
                  LA, USA, December 10 - 16, 2023},
  year         = {2023},
  url          = {http://papers.nips.cc/paper\_files/paper/2023/hash/0764db1151b936aca59249e2c1386101-Abstract-Conference.html},
  bibsource    = {dblp computer science bibliography, https://dblp.org}
}

@article{DBLP:journals/corr/abs-2110-14168,
  author       = {Karl Cobbe and
                  Vineet Kosaraju and
                  Mohammad Bavarian and
                  Mark Chen and
                  Heewoo Jun and
                  Lukasz Kaiser and
                  Matthias Plappert and
                  Jerry Tworek and
                  Jacob Hilton and
                  Reiichiro Nakano and
                  Christopher Hesse and
                  John Schulman},
  title        = {Training Verifiers to Solve Math Word Problems},
  journal      = {CoRR},
  volume       = {abs/2110.14168},
  year         = {2021},
  url          = {https://arxiv.org/abs/2110.14168},
  eprinttype    = {arXiv},
  eprint       = {2110.14168},
  bibsource    = {dblp computer science bibliography, https://dblp.org}
}

@inproceedings{DBLP:conf/naacl/PatelBG21,
  author       = {Arkil Patel and
                  Satwik Bhattamishra and
                  Navin Goyal},
  editor       = {Kristina Toutanova and
                  Anna Rumshisky and
                  Luke Zettlemoyer and
                  Dilek Hakkani{-}T{\"{u}}r and
                  Iz Beltagy and
                  Steven Bethard and
                  Ryan Cotterell and
                  Tanmoy Chakraborty and
                  Yichao Zhou},
  title        = {Are {NLP} Models really able to Solve Simple Math Word Problems?},
  booktitle    = {Proceedings of the 2021 Conference of the North American Chapter of
                  the Association for Computational Linguistics: Human Language Technologies,
                  {NAACL-HLT} 2021, Online, June 6-11, 2021},
  pages        = {2080--2094},
  publisher    = {Association for Computational Linguistics},
  year         = {2021},
  url          = {https://doi.org/10.18653/v1/2021.naacl-main.168},
  doi          = {10.18653/V1/2021.NAACL-MAIN.168},
  bibsource    = {dblp computer science bibliography, https://dblp.org}
}

@article{DBLP:journals/corr/abs-2103-03874,
  author       = {Dan Hendrycks and
                  Collin Burns and
                  Saurav Kadavath and
                  Akul Arora and
                  Steven Basart and
                  Eric Tang and
                  Dawn Song and
                  Jacob Steinhardt},
  title        = {Measuring Mathematical Problem Solving With the {MATH} Dataset},
  journal      = {CoRR},
  volume       = {abs/2103.03874},
  year         = {2021},
  url          = {https://arxiv.org/abs/2103.03874},
  eprinttype    = {arXiv},
  eprint       = {2103.03874},
  bibsource    = {dblp computer science bibliography, https://dblp.org}
}

@article{DBLP:journals/corr/abs-2009-03300,
  author       = {Dan Hendrycks and
                  Collin Burns and
                  Steven Basart and
                  Andy Zou and
                  Mantas Mazeika and
                  Dawn Song and
                  Jacob Steinhardt},
  title        = {Measuring Massive Multitask Language Understanding},
  journal      = {CoRR},
  volume       = {abs/2009.03300},
  year         = {2020},
  url          = {https://arxiv.org/abs/2009.03300},
  eprinttype    = {arXiv},
  eprint       = {2009.03300},
  bibsource    = {dblp computer science bibliography, https://dblp.org}
}

@inproceedings{DBLP:conf/acl/SuzgunSSGTCCLCZ23,
  author       = {Mirac Suzgun and
                  Nathan Scales and
                  Nathanael Sch{\"{a}}rli and
                  Sebastian Gehrmann and
                  Yi Tay and
                  Hyung Won Chung and
                  Aakanksha Chowdhery and
                  Quoc V. Le and
                  Ed H. Chi and
                  Denny Zhou and
                  Jason Wei},
  editor       = {Anna Rogers and
                  Jordan L. Boyd{-}Graber and
                  Naoaki Okazaki},
  title        = {Challenging BIG-Bench Tasks and Whether Chain-of-Thought Can Solve
                  Them},
  booktitle    = {Findings of the Association for Computational Linguistics: {ACL} 2023,
                  Toronto, Canada, July 9-14, 2023},
  pages        = {13003--13051},
  publisher    = {Association for Computational Linguistics},
  year         = {2023},
  url          = {https://doi.org/10.18653/v1/2023.findings-acl.824},
  doi          = {10.18653/V1/2023.FINDINGS-ACL.824},
  bibsource    = {dblp computer science bibliography, https://dblp.org}
}

@article{DBLP:journals/corr/abs-2407-21783,
  author       = {Abhimanyu Dubey and
                  Abhinav Jauhri and
                  Abhinav Pandey and
                  Abhishek Kadian and
                  Ahmad Al{-}Dahle and
                  Aiesha Letman and
                  Akhil Mathur and
                  Alan Schelten and
                  Amy Yang and
                  Angela Fan and
                  Anirudh Goyal and
                  Anthony Hartshorn and
                  Aobo Yang and
                  Archi Mitra and
                  Archie Sravankumar and
                  Artem Korenev and
                  Arthur Hinsvark and
                  Arun Rao and
                  Aston Zhang and
                  Aur{\'{e}}lien Rodriguez and
                  Austen Gregerson and
                  Ava Spataru and
                  Baptiste Rozi{\`{e}}re and
                  Bethany Biron and
                  Binh Tang and
                  Bobbie Chern and
                  Charlotte Caucheteux and
                  Chaya Nayak and
                  Chloe Bi and
                  Chris Marra and
                  Chris McConnell and
                  Christian Keller and
                  Christophe Touret and
                  Chunyang Wu and
                  Corinne Wong and
                  Cristian Canton Ferrer and
                  Cyrus Nikolaidis and
                  Damien Allonsius and
                  Daniel Song and
                  Danielle Pintz and
                  Danny Livshits and
                  David Esiobu and
                  Dhruv Choudhary and
                  Dhruv Mahajan and
                  Diego Garcia{-}Olano and
                  Diego Perino and
                  Dieuwke Hupkes and
                  Egor Lakomkin and
                  Ehab AlBadawy and
                  Elina Lobanova and
                  Emily Dinan and
                  Eric Michael Smith and
                  Filip Radenovic and
                  Frank Zhang and
                  Gabriel Synnaeve and
                  Gabrielle Lee and
                  Georgia Lewis Anderson and
                  Graeme Nail and
                  Gr{\'{e}}goire Mialon and
                  Guan Pang and
                  Guillem Cucurell and
                  Hailey Nguyen and
                  Hannah Korevaar and
                  Hu Xu and
                  Hugo Touvron and
                  Iliyan Zarov and
                  Imanol Arrieta Ibarra and
                  Isabel M. Kloumann and
                  Ishan Misra and
                  Ivan Evtimov and
                  Jade Copet and
                  Jaewon Lee and
                  Jan Geffert and
                  Jana Vranes and
                  Jason Park and
                  Jay Mahadeokar and
                  Jeet Shah and
                  Jelmer van der Linde and
                  Jennifer Billock and
                  Jenny Hong and
                  Jenya Lee and
                  Jeremy Fu and
                  Jianfeng Chi and
                  Jianyu Huang and
                  Jiawen Liu and
                  Jie Wang and
                  Jiecao Yu and
                  Joanna Bitton and
                  Joe Spisak and
                  Jongsoo Park and
                  Joseph Rocca and
                  Joshua Johnstun and
                  Joshua Saxe and
                  Junteng Jia and
                  Kalyan Vasuden Alwala and
                  Kartikeya Upasani and
                  Kate Plawiak and
                  Ke Li and
                  Kenneth Heafield and
                  Kevin Stone and
                  et al.},
  title        = {The Llama 3 Herd of Models},
  journal      = {CoRR},
  volume       = {abs/2407.21783},
  year         = {2024},
  url          = {https://doi.org/10.48550/arXiv.2407.21783},
  doi          = {10.48550/ARXIV.2407.21783},
  eprinttype    = {arXiv},
  eprint       = {2407.21783},
  bibsource    = {dblp computer science bibliography, https://dblp.org}
}

@article{DBLP:journals/corr/abs-2412-15115,
  author       = {An Yang and
                  Baosong Yang and
                  Beichen Zhang and
                  Binyuan Hui and
                  Bo Zheng and
                  Bowen Yu and
                  Chengyuan Li and
                  Dayiheng Liu and
                  Fei Huang and
                  Haoran Wei and
                  Huan Lin and
                  Jian Yang and
                  Jianhong Tu and
                  Jianwei Zhang and
                  Jianxin Yang and
                  Jiaxi Yang and
                  Jingren Zhou and
                  Junyang Lin and
                  Kai Dang and
                  Keming Lu and
                  Keqin Bao and
                  Kexin Yang and
                  Le Yu and
                  Mei Li and
                  Mingfeng Xue and
                  Pei Zhang and
                  Qin Zhu and
                  Rui Men and
                  Runji Lin and
                  Tianhao Li and
                  Tingyu Xia and
                  Xingzhang Ren and
                  Xuancheng Ren and
                  Yang Fan and
                  Yang Su and
                  Yichang Zhang and
                  Yu Wan and
                  Yuqiong Liu and
                  Zeyu Cui and
                  Zhenru Zhang and
                  Zihan Qiu},
  title        = {Qwen2.5 Technical Report},
  journal      = {CoRR},
  volume       = {abs/2412.15115},
  year         = {2024},
  url          = {https://doi.org/10.48550/arXiv.2412.15115},
  doi          = {10.48550/ARXIV.2412.15115},
  eprinttype    = {arXiv},
  eprint       = {2412.15115},
  bibsource    = {dblp computer science bibliography, https://dblp.org}
}

@misc{zeng2024structuralmemoryllmagents,
      title={On the Structural Memory of LLM Agents}, 
      author={Ruihong Zeng and Jinyuan Fang and Siwei Liu and Zaiqiao Meng},
      year={2024},
      eprint={2412.15266},
      archivePrefix={arXiv},
      primaryClass={cs.CL},
      url={https://arxiv.org/abs/2412.15266}, 
}

@misc{li2023largelanguagemodelsunderstand,
      title={Large Language Models Understand and Can be Enhanced by Emotional Stimuli}, 
      author={Cheng Li and Jindong Wang and Yixuan Zhang and Kaijie Zhu and Wenxin Hou and Jianxun Lian and Fang Luo and Qiang Yang and Xing Xie},
      year={2023},
      eprint={2307.11760},
      archivePrefix={arXiv},
      primaryClass={cs.CL},
      url={https://arxiv.org/abs/2307.11760}, 
}

@misc{deepseekai2025deepseekr1incentivizingreasoningcapability,
      title={DeepSeek-R1: Incentivizing Reasoning Capability in LLMs via Reinforcement Learning}, 
      author={DeepSeek-AI and Daya Guo and Dejian Yang and Haowei Zhang and Junxiao Song and Ruoyu Zhang and Runxin Xu and Qihao Zhu and Shirong Ma and Peiyi Wang and Xiao Bi and Xiaokang Zhang and Xingkai Yu and Yu Wu and Z. F. Wu and Zhibin Gou and Zhihong Shao and Zhuoshu Li and Ziyi Gao and Aixin Liu and Bing Xue and Bingxuan Wang and Bochao Wu and Bei Feng and Chengda Lu and Chenggang Zhao and Chengqi Deng and Chenyu Zhang and Chong Ruan and Damai Dai and Deli Chen and Dongjie Ji and Erhang Li and Fangyun Lin and Fucong Dai and Fuli Luo and Guangbo Hao and Guanting Chen and Guowei Li and H. Zhang and Han Bao and Hanwei Xu and Haocheng Wang and Honghui Ding and Huajian Xin and Huazuo Gao and Hui Qu and Hui Li and Jianzhong Guo and Jiashi Li and Jiawei Wang and Jingchang Chen and Jingyang Yuan and Junjie Qiu and Junlong Li and J. L. Cai and Jiaqi Ni and Jian Liang and Jin Chen and Kai Dong and Kai Hu and Kaige Gao and Kang Guan and Kexin Huang and Kuai Yu and Lean Wang and Lecong Zhang and Liang Zhao and Litong Wang and Liyue Zhang and Lei Xu and Leyi Xia and Mingchuan Zhang and Minghua Zhang and Minghui Tang and Meng Li and Miaojun Wang and Mingming Li and Ning Tian and Panpan Huang and Peng Zhang and Qiancheng Wang and Qinyu Chen and Qiushi Du and Ruiqi Ge and Ruisong Zhang and Ruizhe Pan and Runji Wang and R. J. Chen and R. L. Jin and Ruyi Chen and Shanghao Lu and Shangyan Zhou and Shanhuang Chen and Shengfeng Ye and Shiyu Wang and Shuiping Yu and Shunfeng Zhou and Shuting Pan and S. S. Li and Shuang Zhou and Shaoqing Wu and Shengfeng Ye and Tao Yun and Tian Pei and Tianyu Sun and T. Wang and Wangding Zeng and Wanjia Zhao and Wen Liu and Wenfeng Liang and Wenjun Gao and Wenqin Yu and Wentao Zhang and W. L. Xiao and Wei An and Xiaodong Liu and Xiaohan Wang and Xiaokang Chen and Xiaotao Nie and Xin Cheng and Xin Liu and Xin Xie and Xingchao Liu and Xinyu Yang and Xinyuan Li and Xuecheng Su and Xuheng Lin and X. Q. Li and Xiangyue Jin and Xiaojin Shen and Xiaosha Chen and Xiaowen Sun and Xiaoxiang Wang and Xinnan Song and Xinyi Zhou and Xianzu Wang and Xinxia Shan and Y. K. Li and Y. Q. Wang and Y. X. Wei and Yang Zhang and Yanhong Xu and Yao Li and Yao Zhao and Yaofeng Sun and Yaohui Wang and Yi Yu and Yichao Zhang and Yifan Shi and Yiliang Xiong and Ying He and Yishi Piao and Yisong Wang and Yixuan Tan and Yiyang Ma and Yiyuan Liu and Yongqiang Guo and Yuan Ou and Yuduan Wang and Yue Gong and Yuheng Zou and Yujia He and Yunfan Xiong and Yuxiang Luo and Yuxiang You and Yuxuan Liu and Yuyang Zhou and Y. X. Zhu and Yanhong Xu and Yanping Huang and Yaohui Li and Yi Zheng and Yuchen Zhu and Yunxian Ma and Ying Tang and Yukun Zha and Yuting Yan and Z. Z. Ren and Zehui Ren and Zhangli Sha and Zhe Fu and Zhean Xu and Zhenda Xie and Zhengyan Zhang and Zhewen Hao and Zhicheng Ma and Zhigang Yan and Zhiyu Wu and Zihui Gu and Zijia Zhu and Zijun Liu and Zilin Li and Ziwei Xie and Ziyang Song and Zizheng Pan and Zhen Huang and Zhipeng Xu and Zhongyu Zhang and Zhen Zhang},
      year={2025},
      eprint={2501.12948},
      archivePrefix={arXiv},
      primaryClass={cs.CL},
      url={https://arxiv.org/abs/2501.12948}, 
}

@article{rafailov2023direct,
  title={Direct preference optimization: Your language model is secretly a reward model},
  author={Rafailov, Rafael and Sharma, Archit and Mitchell, Eric and Manning, Christopher D and Ermon, Stefano and Finn, Chelsea},
  journal={Advances in neural information processing systems},
  volume={36},
  pages={53728--53741},
  year={2023}
}

@misc{gao2025surveyselfevolvingagentspath,
      title={A Survey of Self-Evolving Agents: On Path to Artificial Super Intelligence}, 
      author={Huan-ang Gao and Jiayi Geng and Wenyue Hua and Mengkang Hu and Xinzhe Juan and Hongzhang Liu and Shilong Liu and Jiahao Qiu and Xuan Qi and Yiran Wu and Hongru Wang and Han Xiao and Yuhang Zhou and Shaokun Zhang and Jiayi Zhang and Jinyu Xiang and Yixiong Fang and Qiwen Zhao and Dongrui Liu and Qihan Ren and Cheng Qian and Zhenghailong Wang and Minda Hu and Huazheng Wang and Qingyun Wu and Heng Ji and Mengdi Wang},
      year={2025},
      eprint={2507.21046},
      archivePrefix={arXiv},
      primaryClass={cs.AI},
      url={https://arxiv.org/abs/2507.21046}, 
}

@article{kim2023cot,
  title={The CoT Collection: Improving Zero-shot and Few-shot Learning of Language Models via Chain-of-Thought Fine-Tuning},
  author={Kim, Seungone and Joo, Se June and Kim, Doyoung and Jang, Joel and Ye, Seonghyeon and Shin, Jamin and Seo, Minjoon},
  journal={arXiv preprint arXiv:2305.14045},
  year={2023}
}

@misc{chen2023selficlzeroshotincontextlearning,
      title={Self-ICL: Zero-Shot In-Context Learning with Self-Generated Demonstrations}, 
      author={Wei-Lin Chen and Cheng-Kuang Wu and Yun-Nung Chen and Hsin-Hsi Chen},
      year={2023},
      eprint={2305.15035},
      archivePrefix={arXiv},
      primaryClass={cs.CL},
      url={https://arxiv.org/abs/2305.15035}, 
}

@misc{zhou2024selfdiscoverlargelanguagemodels,
      title={Self-Discover: Large Language Models Self-Compose Reasoning Structures}, 
      author={Pei Zhou and Jay Pujara and Xiang Ren and Xinyun Chen and Heng-Tze Cheng and Quoc V. Le and Ed H. Chi and Denny Zhou and Swaroop Mishra and Huaixiu Steven Zheng},
      year={2024},
      eprint={2402.03620},
      archivePrefix={arXiv},
      primaryClass={cs.AI},
      url={https://arxiv.org/abs/2402.03620}, 
}

@misc{xu2025amemagenticmemoryllm,
      title={A-MEM: Agentic Memory for LLM Agents}, 
      author={Wujiang Xu and Zujie Liang and Kai Mei and Hang Gao and Juntao Tan and Yongfeng Zhang},
      year={2025},
      eprint={2502.12110},
      archivePrefix={arXiv},
      primaryClass={cs.CL},
      url={https://arxiv.org/abs/2502.12110}, 
}

@misc{huai2025taskcorememorymanagementconsolidation,
      title={Task-Core Memory Management and Consolidation for Long-term Continual Learning}, 
      author={Tianyu Huai and Jie Zhou and Yuxuan Cai and Qin Chen and Wen Wu and Xingjiao Wu and Xipeng Qiu and Liang He},
      year={2025},
      eprint={2505.09952},
      archivePrefix={arXiv},
      primaryClass={cs.LG},
      url={https://arxiv.org/abs/2505.09952}, 
}

@misc{zhong2023memorybankenhancinglargelanguage,
      title={MemoryBank: Enhancing Large Language Models with Long-Term Memory}, 
      author={Wanjun Zhong and Lianghong Guo and Qiqi Gao and He Ye and Yanlin Wang},
      year={2023},
      eprint={2305.10250},
      archivePrefix={arXiv},
      primaryClass={cs.CL},
      url={https://arxiv.org/abs/2305.10250}, 
}

@misc{chhikara2025mem0buildingproductionreadyai,
      title={Mem0: Building Production-Ready AI Agents with Scalable Long-Term Memory}, 
      author={Prateek Chhikara and Dev Khant and Saket Aryan and Taranjeet Singh and Deshraj Yadav},
      year={2025},
      eprint={2504.19413},
      archivePrefix={arXiv},
      primaryClass={cs.CL},
      url={https://arxiv.org/abs/2504.19413}, 
}

@misc{liu2025contextualexperiencereplayselfimprovement,
      title={Contextual Experience Replay for Self-Improvement of Language Agents}, 
      author={Yitao Liu and Chenglei Si and Karthik Narasimhan and Shunyu Yao},
      year={2025},
      eprint={2506.06698},
      archivePrefix={arXiv},
      primaryClass={cs.AI},
      url={https://arxiv.org/abs/2506.06698}, 
}

@misc{ouyang2025reasoningbankscalingagentselfevolving,
      title={ReasoningBank: Scaling Agent Self-Evolving with Reasoning Memory}, 
      author={Siru Ouyang and Jun Yan and I-Hung Hsu and Yanfei Chen and Ke Jiang and Zifeng Wang and Rujun Han and Long T. Le and Samira Daruki and Xiangru Tang and Vishy Tirumalashetty and George Lee and Mahsan Rofouei and Hangfei Lin and Jiawei Han and Chen-Yu Lee and Tomas Pfister},
      year={2025},
      eprint={2509.25140},
      archivePrefix={arXiv},
      primaryClass={cs.AI},
      url={https://arxiv.org/abs/2509.25140}, 
}

@misc{xu2025sedmscalableselfevolvingdistributed,
      title={SEDM: Scalable Self-Evolving Distributed Memory for Agents}, 
      author={Haoran Xu and Jiacong Hu and Ke Zhang and Lei Yu and Yuxin Tang and Xinyuan Song and Yiqun Duan and Lynn Ai and Bill Shi},
      year={2025},
      eprint={2509.09498},
      archivePrefix={arXiv},
      primaryClass={cs.AI},
      url={https://arxiv.org/abs/2509.09498}, 
}

@misc{li2024robocoderroboticlearningbasic,
      title={RoboCoder: Robotic Learning from Basic Skills to General Tasks with Large Language Models}, 
      author={Jingyao Li and Pengguang Chen and Sitong Wu and Chuanyang Zheng and Hong Xu and Jiaya Jia},
      year={2024},
      eprint={2406.03757},
      archivePrefix={arXiv},
      primaryClass={cs.RO},
      url={https://arxiv.org/abs/2406.03757}, 
}

@misc{cai2025buildingselfevolvingagentsexperiencedriven,
      title={Building Self-Evolving Agents via Experience-Driven Lifelong Learning: A Framework and Benchmark}, 
      author={Yuxuan Cai and Yipeng Hao and Jie Zhou and Hang Yan and Zhikai Lei and Rui Zhen and Zhenhua Han and Yutao Yang and Junsong Li and Qianjun Pan and Tianyu Huai and Qin Chen and Xin Li and Kai Chen and Bo Zhang and Xipeng Qiu and Liang He},
      year={2025},
      eprint={2508.19005},
      archivePrefix={arXiv},
      primaryClass={cs.AI},
      url={https://arxiv.org/abs/2508.19005}, 
}

\appendix

\end{document}